# Using Machine Learning and Neural Networks to Analyze and Predict Chaos in Multi-Pendulum and Chaotic Systems

Vasista Ramachandruni[a], Sai Hruday Reddy Nara[a], Geo Lalu[a], Sabrina Yang[a], Mohit Ramesh Kumar[a], Aarjav Jain[a], Pratham Mehta[a], Hankyu Koo[a], Jason Damonte[a], Marx Akl[a]**

[a]*Department of Computer Science and Engineering, Aspiring Scholars Directed Research Program, 46249 Warm Springs Blvd., Fremont, California, 94539, USA*
**Corresponding author*

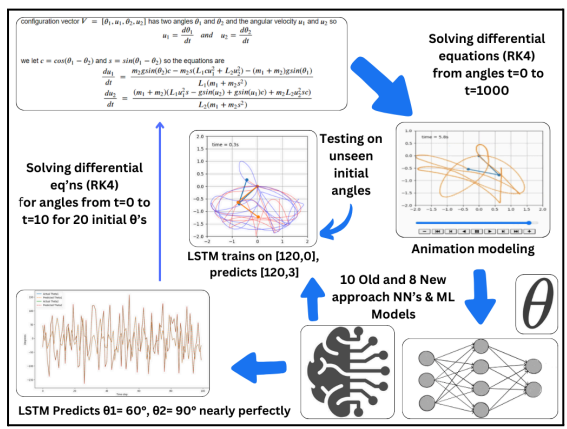

## ABSTRACT

A chaotic system is a highly volatile system characterized by its sensitive dependence on initial conditions and outside factors. Chaotic systems are prevalent throughout the world today: in weather patterns, disease outbreaks, and even financial markets. Chaotic systems are seen in every field of science and humanities, so being able to predict these systems is greatly beneficial to society. In this study, we evaluate 10 different machine learning models and neural networks [1] based on Root Mean Squared Error (RMSE) and $R^2$ values for their ability to predict one of these systems, the multi-pendulum. We begin by generating synthetic data representing the angles of the pendulum over time using the Runge Kutta Method for solving 4th Order

Differential Equations (ODE-RK4) [2]. At first, we used the single-step sliding window approach, predicting the 50st step after training for steps 0-49 and so forth. However, to more accurately cover chaotic motion and behavior in these systems, we transitioned to a time-step based approach. Here, we trained the model/network on many initial angles and tested it on a completely new set of initial angles, or "in-between" to capture chaotic motion to its fullest extent. We also evaluated the stability of the system using Lyapunov exponents. We concluded that for a double pendulum, the best model was the Long Short Term Memory Network (LSTM)[3] for the sliding window and time step approaches in both friction and frictionless scenarios. For triple pendulum, the Vanilla Recurrent Neural Network (VRNN)[4] was the best for the sliding window and Gated Recurrent Network (GRU) [5] was the best for the time step approach, but for friction, LSTM was the best.

# 1. INTRODUCTION

Chaotic systems [6] are prevalent in various natural and engineered environments, from weather systems[7] and ecosystems to biomedicine [8] and financial markets[9]. These systems, characterized by their sensitivity to initial conditions[10], often exhibit unpredictable and complex behavior. Understanding and predicting the dynamics of chaotic systems[11] is crucial for better decision-making and control in engineering, economics, and meteorology, minimizing risks and enhancing performance. For example, the accurate prediction of the chaotic system of winds and weather could allow us to plan more efficient flight routes at more efficient times to minimize fuel usage and minimize travel time[12]. In addition, it could aid our ability to predict potential catastrophic weather events. Our research focuses on the multi-pendulum system, the simplest chaotic system. By being able to predict the multi-pendulum system, we can use insights from the machine learning models and neural networks [13] detailed in this paper to create machine learning models and neural networks that can predict various chaotic systems well using minimal resources. Furthermore, this research would also indicate models and/or neural networks that predict chaotic systems such as the multi-pendulum [14] the best, reducing time and resources needed to evaluate various machine learning techniques and neural networks.

The multi-pendulum system, specifically the double pendulum [15] consists of one pendulum attached to an immovable point, and a second pendulum attached to the end of the first pendulum {Figure 1}. For the triple pendulum [16], a third pendulum is attached to the second pendulum {Figure 1}. The multi-pendulum system is a simple yet powerful model that illustrates the complex dynamics arising from non-linear interactions. Despite its simplicity, predicting the future states of the multi-pendulum accurately is challenging due to its highly sensitive dependence on initial conditions and the intricate interplay between its components.

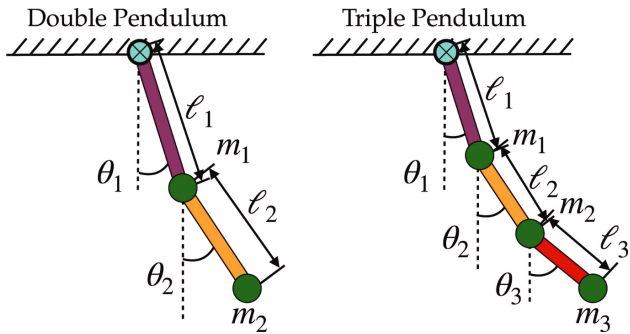
**Fig 1.** An illustration of the double and triple pendulum with angles, masses, and lengths of the pendulums labeled. [17]

In relatively simple classical mechanics problems such as the multi-pendulum system, the position of the system is given by differential equations [17] {Figure 3}. To get the trajectory of a multi-pendulum system, we first need to solve the differential equations of motion. The double pendulum consists of two second-order differential equations.

Configuration vector $V = [\theta_1, u_1, \theta_2, u_2]$ has two angles $\theta_1$ and $\theta_2$ and the angular velocity $u_1$ and $u_2$ defined as:

$u_1 = d\theta_1/dt$ and $u_2 = d\theta_2/dt$

We let $c = \cos(\theta_1 - \theta_2)$ and $s = \sin(\theta_1 - \theta_2)$ so the equations are:

$$du1/dt = \frac{m_2 g \sin(\theta_2)c - m_2 s (L_1 u_1^2 + L_2 u_2^2 c) - (m_1 + m_2)g \sin(\theta_1)}{L(m_1 + m_2 s^2)} \quad (1)$$

$$du2/dt = \frac{(m_1 + m_2)(L_1 u_1^2 s - g\sin(\theta_1) + g \sin(\theta_2)c) + m_2 L_2 u_2^2 sc}{L_2 (m_1 + m_2 s^2)]} \quad (2)$$

(Proof of equations / editable equations in [Electronic Supplementary Information](#))

For triple pendulum, the equations are:

$$du1/dt = \frac{mf(od1\_1 \cdot od1\_2 \cdot od1\_3 \cdot r2 - 4 \cdot (-m3 \cdot od1\_2 \cdot r1 + (m3 \cdot r12 + m12 \cdot r3 \cdot r2) \cdot \cos(\theta1 - \theta2)) \cdot od1\_4 - (od1\_5 + 4m3r12 + 4m12r3r2) \cdot od1\_6}{l1 \cdot od1\_7 \cdot m012 \cdot r2}$$

$$du2/dt = \frac{-m3 \cdot r1 \cdot m012 \cdot od2\_1 \cdot r2 - (m3 \cdot (r1\cos(\theta1 - \theta2) + r2\cos(\theta1 - \theta3)) \cdot r1 - (m3 \cdot r12 + m12 \cdot r3 \cdot r2) \cdot \cos(\theta1 - \theta2)) \cdot od2\_2 + m012 \cdot r3 \cdot r2 \cdot od2\_3}{l2 \cdot od1\_7 \cdot r2}$$

$$du3/dt = -\frac{m12 \cdot (od1\_2) \cdot (od3\_1) + m12 \cdot m012 \cdot (od3\_2) \cdot r2 - r1 \cdot m012 \cdot od3\_3}{l3 \cdot (m3 \cdot r12 + m12 \cdot r3 \cdot r2)}$$

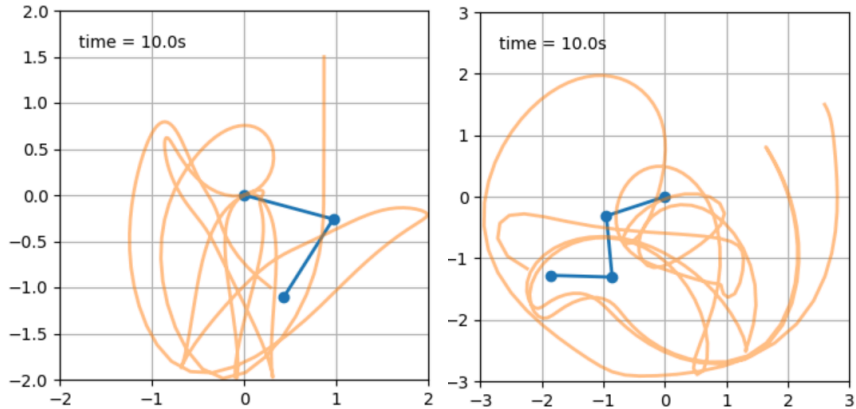

**Fig 2 (up).** Animation of the double and triple pendulum. 10s of animation. This was created by us using the differential equations from cloud4science et al. [7] for the double pendulum, and Yesilyurt [9] for the triple pendulum. Immovable point at (0,0), each pendulum is of length one. Chaotic motion is clearly presented.

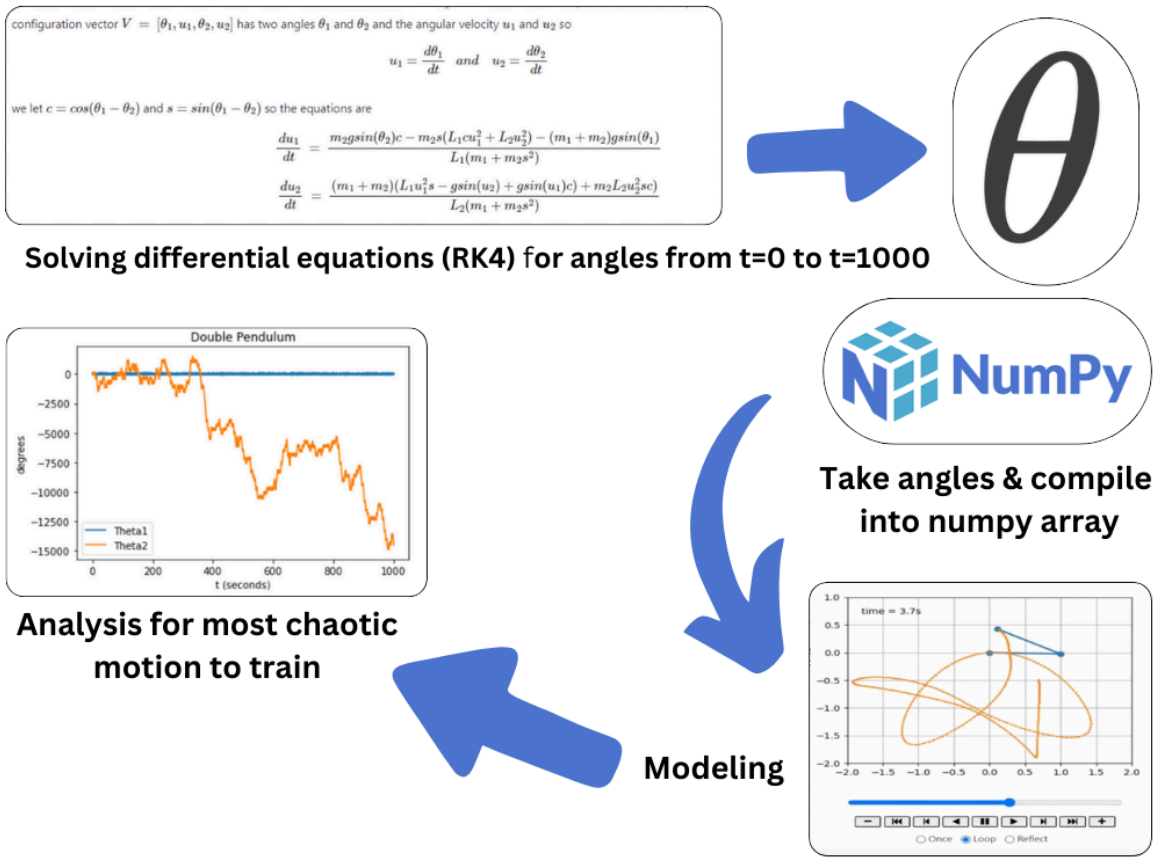

**Fig 3.** Data Preprocessing Workflow

We begin by solving the differential equations for the double pendulum for angles [18] from timestep 0 to 1000, with a step size of 0.001. We take the synthetic data and save it in a 2D NumPy array [19]: one array for $\theta_1$ and one array for $\theta_2$. We then modeled the pendulum through MatPlotLib [20]. We then analyzed the different pendulum data trajectories to determine the balance between predictability and chaos. Further details can be found in the Electronic Supplementary Information.

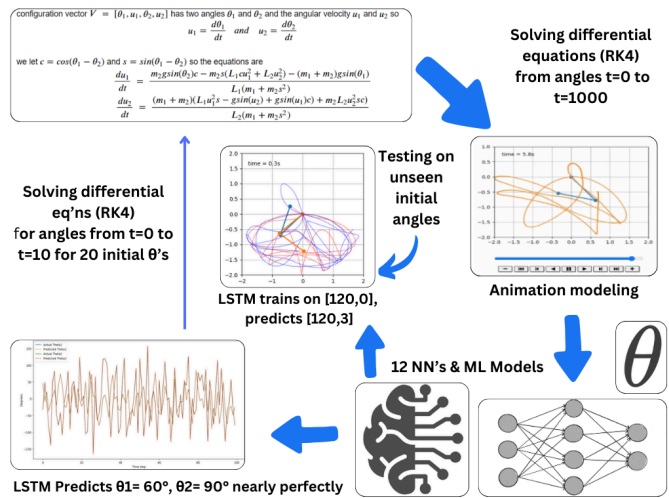

**Fig. 4.** Workflow for model testing.

Using the 10,000,000 synthetic data points for the initial angles [90°, 90°] where the angles are in the format [$\theta_1$, $\theta_2$] created for training and testing data, we initially formed one comprehensive dataset to use for training and testing. Our features initially included the two angle arrays, $\theta_1$ and $\theta_2$, which contain angle measurements at each of the 10,000,000 timesteps.

In our study, we employ both machine learning and neural network techniques to analyze and predict the behavior of the double and triple pendulum. Our method leverages the power of sequential and supervised neural networks and models to get more accurate predictions. Our work differs from previous research in several significant ways. One of the most similar projects, <u>Evaluating Current Machine Learning Techniques On Predicting Chaotic Systems</u> by S. Klinkachorn and J. Parmar from Stanford University [21], utilizes 4 of the same models, but uses a different approach in predictions. It involves minimal analysis of the double pendulum itself and no analysis of the triple pendulum. As such, our approach, while inspired by such methodologies, is generalized for multi-pendulum systems as a whole, showcasing its broader applicability in chaotic systems analysis. Among the models tested in this study, Long-Short Term Memory (LSTM) network and Autoregressive (AR) Model presented the best results; LSTM achieving an RMSE of $1.4_E^{-2}$ and an $R^2$ score of 0.998, and AR achieving an RMSE of $1.5_E^{-2}$ and an $R^2$ score of 0.998. From those results, we can successfully showcase the power of machine learning and deep learning in the scope of predicting chaos, and its application in the real world. Along with neural networks to assess the multi-pendulum system, we employed both the Lyapunov exponent and eigenvalues for the double pendulum differential equations to quantify the stability of the system at various initial conditions and equilibrium points.

## 2. METHODS AND MATERIALS

First, using the equations for the double pendulum the Runge-Kutta 4th Order Method (ODE-RK4), and these initial conditions {Figure 5A} we calculated the time evolution of the pendulum angles from 0 to 1000 seconds with a step size of 0.001. We then took the angle at each step as calculated by the differential equations and pushed the data points into a NumPy array. These 10,000,000 synthetic data points were then scaled using MinMaxScaler and normalized to ensure effective model training.

**Initial conditions:**
$g = 9.81$ m/s²
$l_1 = 1$ m
$l_2 = 1$ m
$m_1 = 1$ kg
$m_1 = 1$ kg

Key - g: Gravitational Constant on Earth ($\frac{m}{s^2}$), L: Length of Pendulum 1/2 (meters), m: mass of Pendulum 1/2 (kg), y: position (given recursively), k: Approximations of the derivative at different points within our iteration, h: Size of each step

**Fig. 5:** A: Initial Conditions and B: ODE-RK4 Algorithm

After solving the differential equations with ODE-RK4, we created an animation using Matplotlib to plot the double and triple pendulum data points at a steep angle ([120,120]) to visualize the trajectory graph over a 10-second interval. {Figure 2A, 2B}.

Due to high sensitivity to initial conditions {Figure 6}, we tested different initial conditions to see how they would affect both trajectories of the double pendulum.

6A:

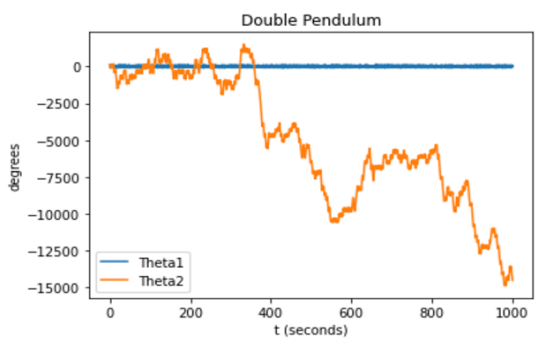

6B:

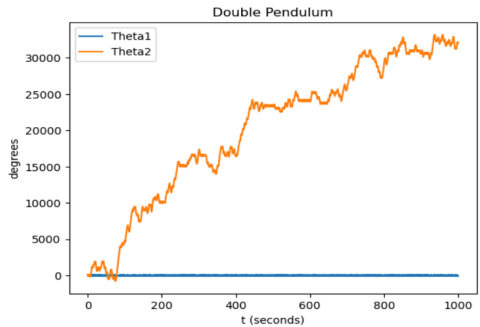

**Fig. 6: RK4 Result Graph** Solved differential equations, showing the different angles for θ$_1$ and θ$_2$ over the time interval (0, 100). Initial angles for this graph: [90°, 90°] vs. RK4 Result Graph/ solved differential equations, showing the different angles for θ$_1$ and θ$_2$ over the time interval (0, 1000). <u>Initial conditions for 6A</u>: gravity: 9.81 $\frac{m}{s^2}$ (standard gravity), $l_1$: 1m, $l_2$: 2m, m$_1$: 1kg, $m_2$: 1kg, θ$_1$: 60°, θ$_2$: 90°. <u>Initial conditions for 6B</u>: gravity: 9.81 $\frac{m}{s^2}$ (standard gravity), $l_1$: 1m, $l_2$: 2m, ==$m_1$: 2kg,== $m_2$: 1kg, θ$_1$: 60°, θ$_2$: 90°.
==*Changed condition*==

The difference between the graphs of Figure 6 exemplifies the influence that initial conditions have on the double pendulum. For simplicity and to generate a good baseline, we decided to stick to modifying only initial angles without considering friction instead of other variables of interest such as gravity, mass, or length. We later moved on to include friction in our analysis as well. We then analyzed the pendulums' graphs and trajectories to determine patterns of chaos to gain a deeper understanding of the system. One seemingly apparent pattern would be the consistency of θ$_1$, but this pattern stems from the fact that the first pendulum goes through roughly the same angles. (citation needed)

Initially, we considered 1000 seconds of data as sufficient enough to model the chaotic double pendulum as this would adequately cover the true chaos in the system past the initial period of stability seen. Our goal is to be able to predict the motion of the pendulum at its most chaotic stages and see how models perform with as minimum data [22] as possible. Chaos develops in the double pendulum gradually. Because this data is minimal, we used Adam Optimizer [23] – a learning rate regulator – to prevent overfitting [24], with a minimum learning rate of 10$^{-4}$.

The initial dataset ingested into each model tested in this study had a size of 10,000,000 points, each representing an angle measure of the pendulum at a specific time step in that 1000 second interval. The feature of the model was a stacked NumPy array of the angles [θ$_1$, θ$_2$] created by ODE-RK4's solution to the differential equation (More information about the approach can be found in [Electronic Supplementary Information](#)).

The following machine learning models and neural networks were trained and evaluated for this study: Random Forest Regressor (RF) [25], Multi-Layer Perceptron Regressor (MLPRegressor) [26], Long-Short Term Memory Neural Network (LSTM), Vanilla Recurrent

Neural Network(VRNN), BiDirectionalRNN (BIRNN) [27], StackedRNN(STRNN), SGDRegressor (SGD) [28], Gated Recurrent Network (GRU), Autoregressive Model (AR) [29], and Feed Forward Neural Network (FFNN). All models were implemented with Python's SciKit-Learn module. PyTorch was used to convert data into tensors for all neural networks [30]. Our results were evaluated using two metrics: $R^2$ score and Root Mean Squared Error (RMSE). We did this as each metric has its own purpose: $R^2$ is a numerical representation of the data's fit to the model, whereas RMSE is a numerical representation of how well the model is able to extrapolate the data given. We consider RMSE as the preferred analysis metric as it provides insight on a model's ability to make predictions, rather than its applicability to the double pendulum model specifically.

Triple pendulum analysis consisted of the same steps, except using differential equations pulled from Yesilyurt and input sequences being a stacked NumPy array with angles [$\theta_1$, $\theta_2$, $\theta_3$]. ODE-RK4 was used in the same way as for the double pendulum. Initial conditions were the same for each of the three pendulums, specifically [90°, 90°, 90°] with standard mass, gravity, and length. This is similar to the double pendulum where initial conditions [90°, 90°] for each of the two pendulums, with the same standard unchanged conditions. The exact same models were trained and analyzed with the same metrics and analysis practices as with the double pendulum.

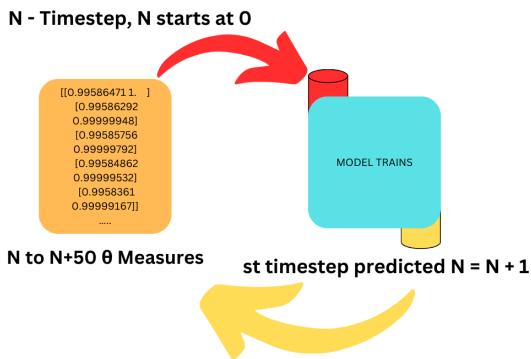

**Fig. 7 Neural Network Sliding Window Approach Model Architecture** Angle measures of the double and triple pendulums at the first 50 time steps were taken. The model trains on these data points and predicts the 51st data point.

## 2.1 Sliding Window

To evaluate the neural network, we used a single-step sliding window approach for predictions [31]. In this approach, sequences of 49 inputs (one input is a set of angles); and predicts one step (one set of angles at the consequent timestep). In the first iteration, angles from timestep 0 to timestep 49 are part of the sequence of inputs, and the model predicts the 50th. In the second iteration, angles from timestep 1 to 50 from the actual data are included in the input sequence. [32] From this, the model predicts the 51st step. This process continues over the 10,000,000 data points in our training set. The sliding window approach is widely used in time series problems such as the double pendulum, as it makes it easier for the model to capture temporal dependencies in the data. [33]

This sliding window approach is good for short term predictions confined to a specific set of initial angles, but not necessarily for multiple different initial angles. [34] In fact, chaotic systems, by definition, are characterized by their sensitive dependence on initial conditions [35], and at this point, initial conditions weren't being changed in our project and true total chaos wasn't being predicted; rather, we were just fitting a very erratic curve. In addition, as we were using the first 49 data points for training and never testing on the same data points, we were not able to construct an accurately visualized trajectory. As a result, we started to work backwards to come up with a new approach that would address these three issues.

## 2.2 Time-step Based Approach

After brainstorming, we came up with the idea of training our best models on many different certain initial angles with a timestep based approach, and then testing it on a completely new untrained set of initial angles within the range of initial angles tested. For example, if our model was trained on initial angles [120°, 0], [120°, 0.1°]...[120°, 2.9]...[120°, 3.0°], we would pick an initial angle such as [120°, 2.05°] and ask our model to predict the trajectory. In addition, this approach doesn't use a sliding window, but an input feature, the timestep: the time at which each angle is observed in the double pendulum system, which allows the model to predict for a specific point in time, allowing us to easily visualize and compare it with [36]. Lastly, the approach takes in the initial angle in every timestep input, allowing the model to differentiate the initial models and come up with its own way to predict previously untrained "inbetweens", or untrained initial angles, whose trajectories chaotically differ from the trained trajectories, allowing us to truly–to a certain extent–predict the trajectory of chaotic systems [37]. This approach also allows us to fully predict the pendulum from 0s to 10s, allowing us to make a visualized trajectory graph comparison,and fully showing the capability of the model to predict the entire interval of chaos. This approach fully captures the chaos of the double pendulum system and allows for patterns to be recognized by these models to predict unknown systems as we train multiple different types of models using the same approach and preprocessing [38]. This allows us to see which type of machine learning model or neural network proves most promising at generalizing and predicting chaotic systems as a whole. **By creating the novel time-step approach, we were able to successfully predict true chaotic motion, and showcase the power of machine learning and artificial intelligence in solving those problems.**

Again, by using ODE-RK4, we generated synthetic data for the double pendulum. However this time only over 10 seconds for each initial angle. We had to figure out the most optimal angle for chaos, as we reduced the training window to 10s for faster computing time of multiple different initial angles instead of only one. We trained on 25 initial angles including [120°, 0], [120°, 0.1°] … [120°, 1.0°], [120°, 1.1] … [120°, 2.0°] ...[120°, 3.0°]. These were all combined into a stacked Numpy array, but with further subdivisions for each initial angle to ensure training on each initial angle separately.

The final dataset contained 2,000 data points for each initial angle, of which there were thirty trained on. In total, our dataset consisted of 60,000 unique data points to train and test our models on, and an extra 2,000 data points for testing on the "inbetween" of [120°, 2.05°].

The same models were trained for the time-step based approach as the sliding window approach, with the exception of SGD Regressor and Random Forest. We then compared the visualized trajectories for each model prediction, the graph between predicted, the actual angles, and the RMSE to determine the best model for predicting the chaotic motion of the multi-pendulum system.

We then created a program to test our model on completely new conditions (eg. [120°, 2.05°]), and evaluate how accurate the prediction compares to the actual values. Training on these completely new conditions means that the model has to adapt to a pattern it has never recognized before. A slight tweak in initial angles drastically changes the time-angle relation of the pendulum, making it almost "unpredictable." Based on our results, we were able to verify which models performed the best in our previous approach, and also examined which models would be able to predict based on completely unseen data the best. Our results were again evaluated using two metrics: $R^2$ score and Root Mean Squared Error (RMSE). We also compared the visualized trajectories for each model prediction again to see how our models performed in relation to each other, and in relation to the sliding window approach.

In order to understand the dynamics of the double pendulum further, we conducted stability analysis to find the non-chaotic points of the system, using the Lyapunov exponent [18] with the differential equations by finding the slope of the regression line of the data: $(t, ln(error(t)))$ for 10000 timesteps at every starting condition $\theta_1$, $\theta_2 \in [0, 180]$ with increments of 15° and $\theta_1$ perturbation of 0.1°. The error at $t$ was calculated with $error(t) = |\theta_{perturbed} - \theta_{actual}|$. Using this, we were able to find relatively non-chaotic initial conditions of the double pendulum system.

## 3. RESULTS AND DISCUSSION

### 3.1 Angle and Trajectory Analysis

Many groups, such as Chattopadhyay et. al (2020) and Goodfellow et al. (2016), focus on using deep neural networks for predicting the dynamics of chaotic systems [39][40]. These studies typically take advantage of single-layer or shallow neural networks for modeling the dynamics of the multi-pendulum system [41]. Our first result was our analysis of the double pendulum solutions [42]. It revealed that when disregarding outside factors such as friction or air resistance, there were certain patterns in terms of chaotic motion based on the initial angles of the pendulum. Specifically for the double pendulum, when [θ₁,θ₂] ≥ [89.1089°,59.6026°], the pendulum demonstrated extremely chaotic motion, fully rotating around the origin in multiple instances over time {Figure 8A}. However, if [θ₁,θ₂] ≥ [59.6026°, 89.1089°], chaos is a lot less prevalent; the trajectory almost resembles what a single pendulum might look like {Figure 8B}. From this data, we hypothesized that an increase in $\theta_1$ has a greater correlation with chaos in the pendulum than $\theta_2$, verifying this through initial angles that is by testing various different initial angles.

8A:

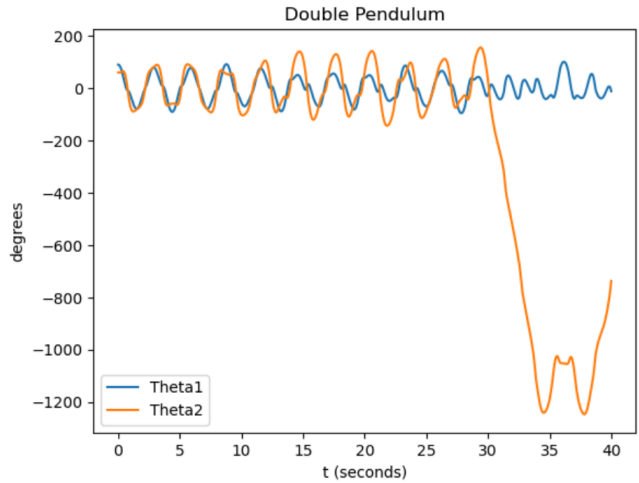

8B:

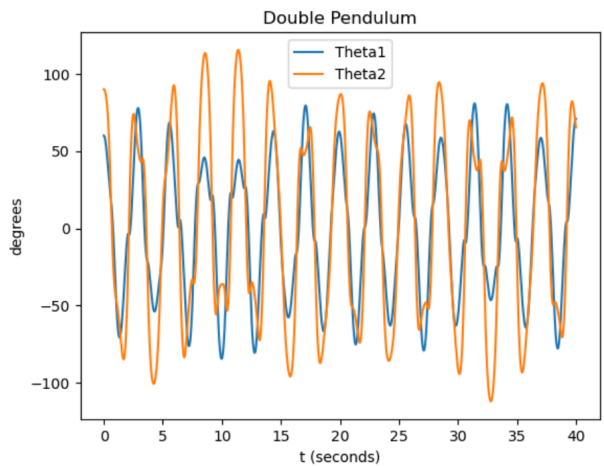

8C:

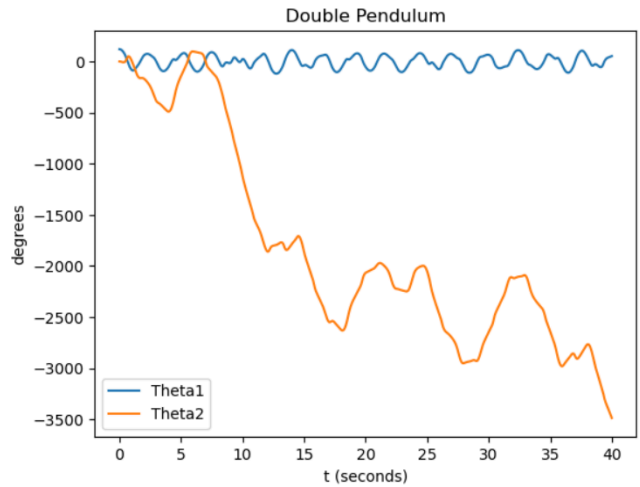

8D:

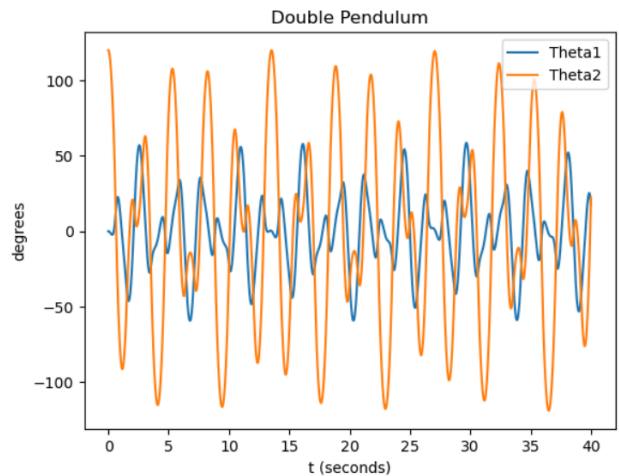

**Fig 8A, 8B, 8C, 8D. Relative Chaotic vs. Non-Chaotic Motion: Frictionless**; Figure 8A depicts the ODE-RK4 solution graph with initial angles $[\theta_1,\theta_2]$ = [90°,60°]; Figure 8B depicts the ODE-RK4 solution graph with initial angles $[\theta_1,\theta_2]$ = [60°, 90°]. These graphs highlight the border point between chaos and stability. Figure 8C depicts the ODE-RK4 solution graph with initial angles $[\theta_1,\theta_2]$ = [120°, 0°]. This demonstrates almost unpredictable chaos. By changing the initial angles to $[\theta_1,\theta_2]$ = [0°, 120°], we see a significant decrease in chaotic behavior {Figure 8D}. A, B, C, D all suggest that $\theta_1$ affects chaotic motion in a frictionless system to a greater magnitude than $\theta_2$.

9A:

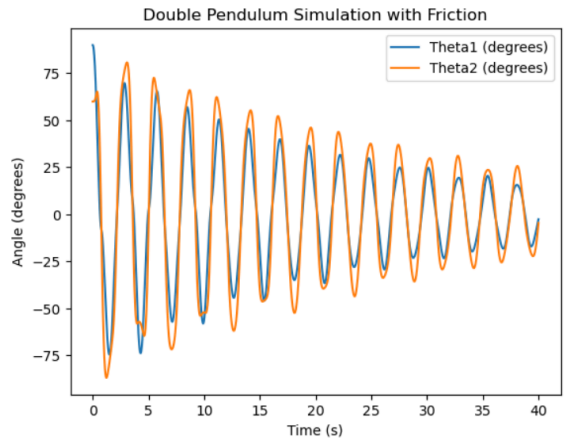

9B:

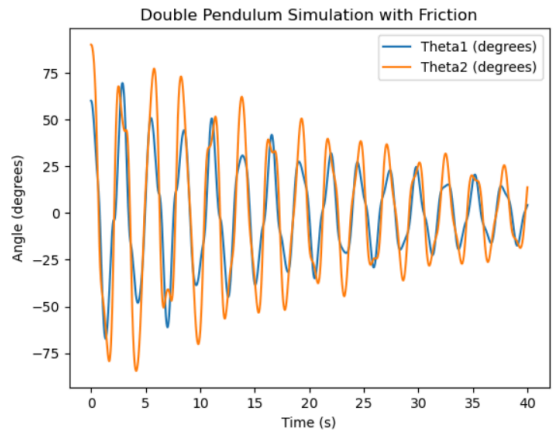

9C:

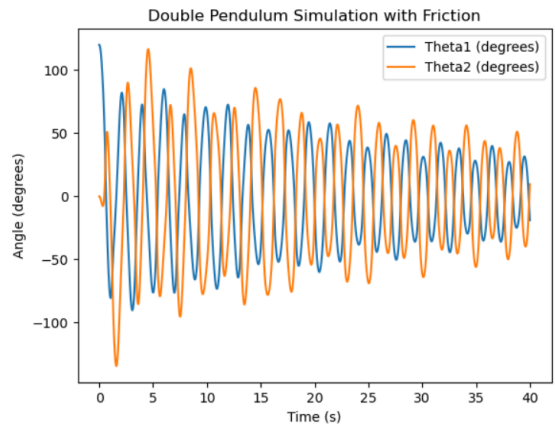

9D:

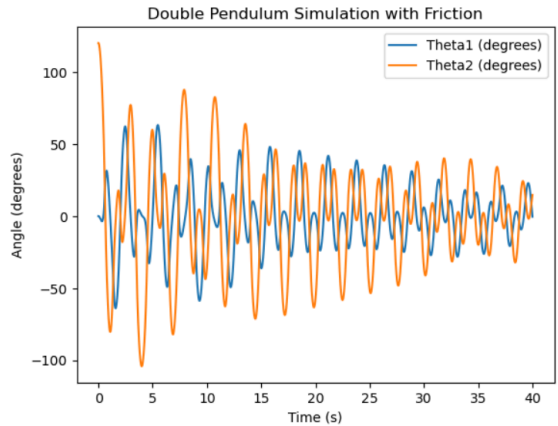

**Fig 9A, 9B, 9C, 9D. Relative Chaotic vs. Non-Chaotic Motion In Friction**; Figure 9A depicts the ODE-RK4 solution graph with initial angles [$\theta_1,\theta_2$] = [90°,60°]; Figure 9B depicts the ODE-RK4 solution graph with initial angles [$\theta_1,\theta_2$] ≥ [60°, 90°]. Figure 9C depicts the ODE-RK4 solution graph with initial angles [$\theta_1,\theta_2$] = [120°, 0°]. Figure 9D depicts the ODE-RK4 solution graph with initial angles [$\theta_1,\theta_2$] ≥ [0°, 120°]. All solutions were derived with the application of friction into the differential equations.

The application of friction into the differential equations was done through the use of a damping constant, *damping1* for the first pendulum, and *damping2* for the second pendulum. By analyzing the initial angles and their solutions, we were able to demonstrate that the opposite trend occurred for friction: instead of $\theta_1$ having greater impact on the chaos of the system, $\theta_2$ had a greater impact, shown by the stable pendulum motion in Figure 9A and Figure 9C and the slightly jagged turns at the highest potential energy points of the pendulum in Figure 9B and Figure 9D. In addition, larger angles, as expected, demonstrated the greatest chaos for friction, with figures 9A and 9B demonstrating limited chaotic motion compared to Figures 9C and 9D.

From these graphs, we initially concluded that training our first models with the sliding window approach at initial angle [90°, 90°] without friction would be suitable for the double pendulum, as it demonstrated the best balance between chaos and predictability for a large time array (0s to 1000s). Therefore, we based our initial sliding window approach models entirely on the basis of this analysis. However, after realizing that the sliding window approach did not model chaotic motion and after creating the time-step approach, we had to change our time interval and the magnitude of the initial angle. In the sliding window approach, training and testing the model on a 10,000,000 point dataset was extremely time-consuming on our machines with limited computing capability. Training on multiple angles, like in the time-step based approach, with 1000s seconds of motion for each angle, would cost too much computational power and time, so we had to decrease the interval. However, decreasing the time frame allows for less chaos to develop in the system, so we had to increase our initial angle magnitude as well. Our analysis of angle [120°, 0°] led us to believe that this angle would develop the greatest chaos in the smaller time frame, and would still be predictable by the models. For the multiple different initial angles, we found that incrementing θ₂ by 0.1° gave us the most predictable deviation from the previous angle. Our final choice of angles for the time-step approach was all angles from [120°, 0°] to [120°, 3°], with increments of 0.1° in θ₂, giving us 30 initial angles to train on, or 60,000 unique data points for training. We could then test on an "in-between" (e.g. [120°, 2.05°]) and see how the models would perform on new unseen time-series data. Here, we applied both friction-based and frictionless scenarios with the same initial angles to train on, comparing model performance on relatively non-chaotic motion with friction opposed to highly chaotic motion without friction.

## 3.2 Double Pendulum

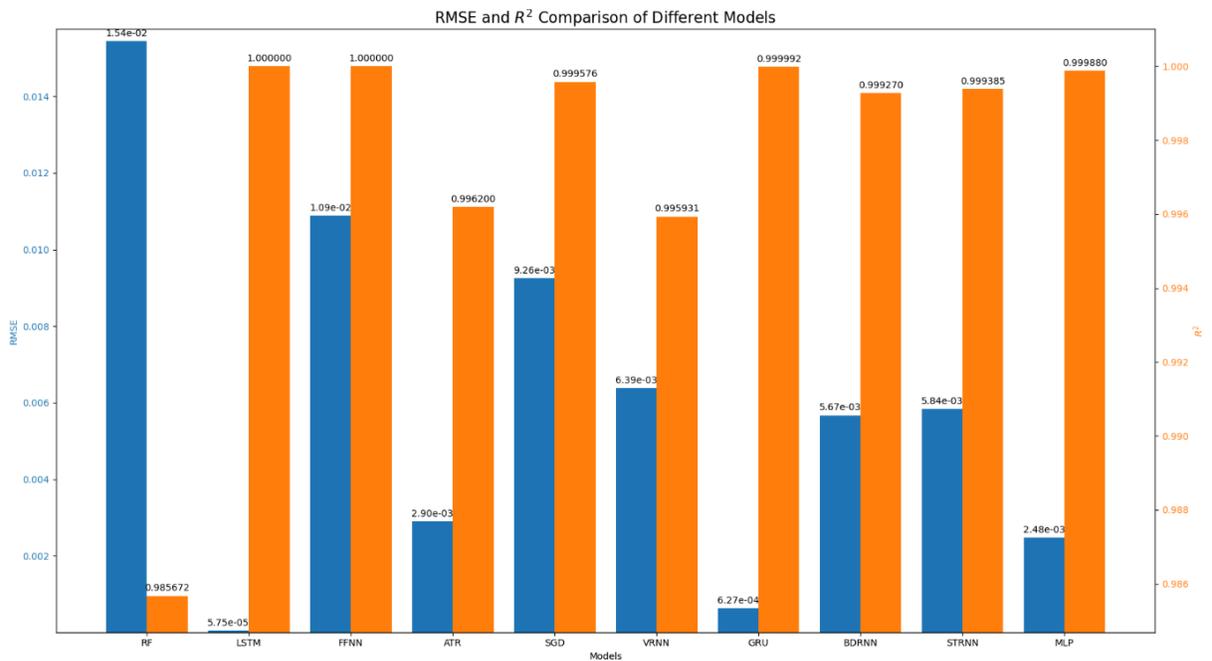

**Fig. 10: Double Pendulum Sliding Window Model Results** Performances on various ML and NN algorithms on the dataset. Evaluation metrics used were $R^2$ and RMSE for initial angle [90°, 90°]. All results did not take the aspect of friction into consideration.

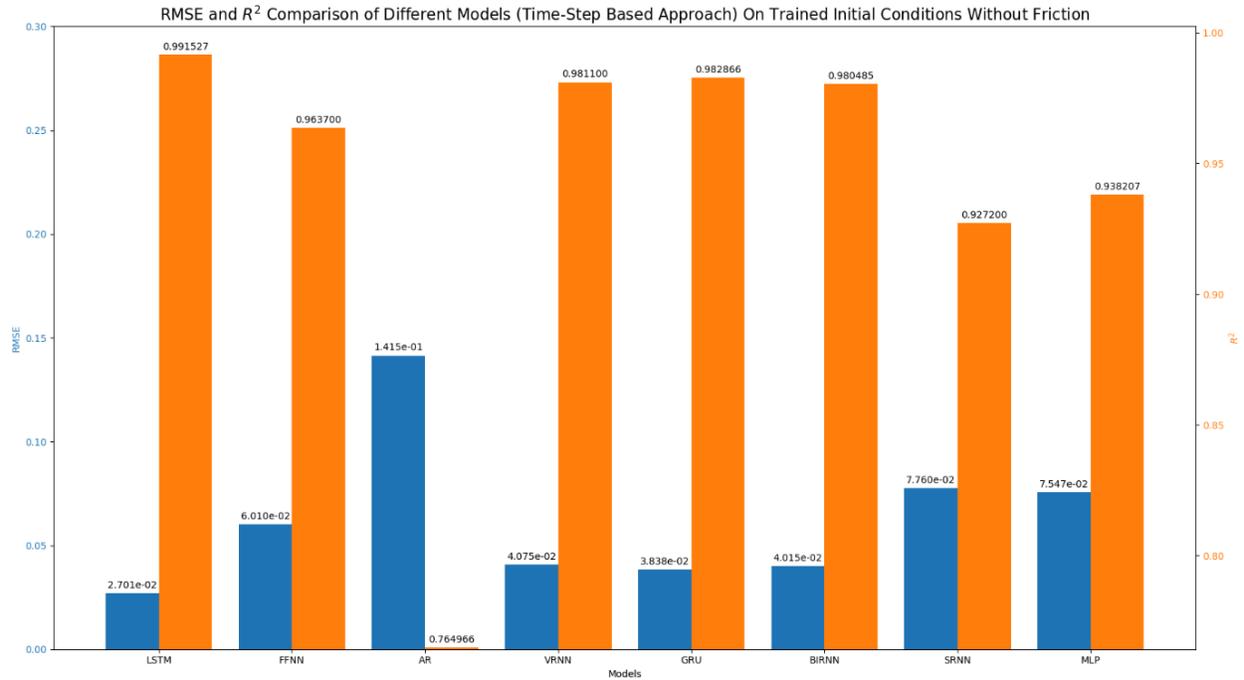

**Fig. 11: Double Pendulum Time-Step No Friction Model Results** Performances on various ML and NN algorithms on the dataset. Evaluation metrics used were $R^2$ and RMSE for initial angle [120°, 0°]. All results did not take the aspect of friction into consideration.

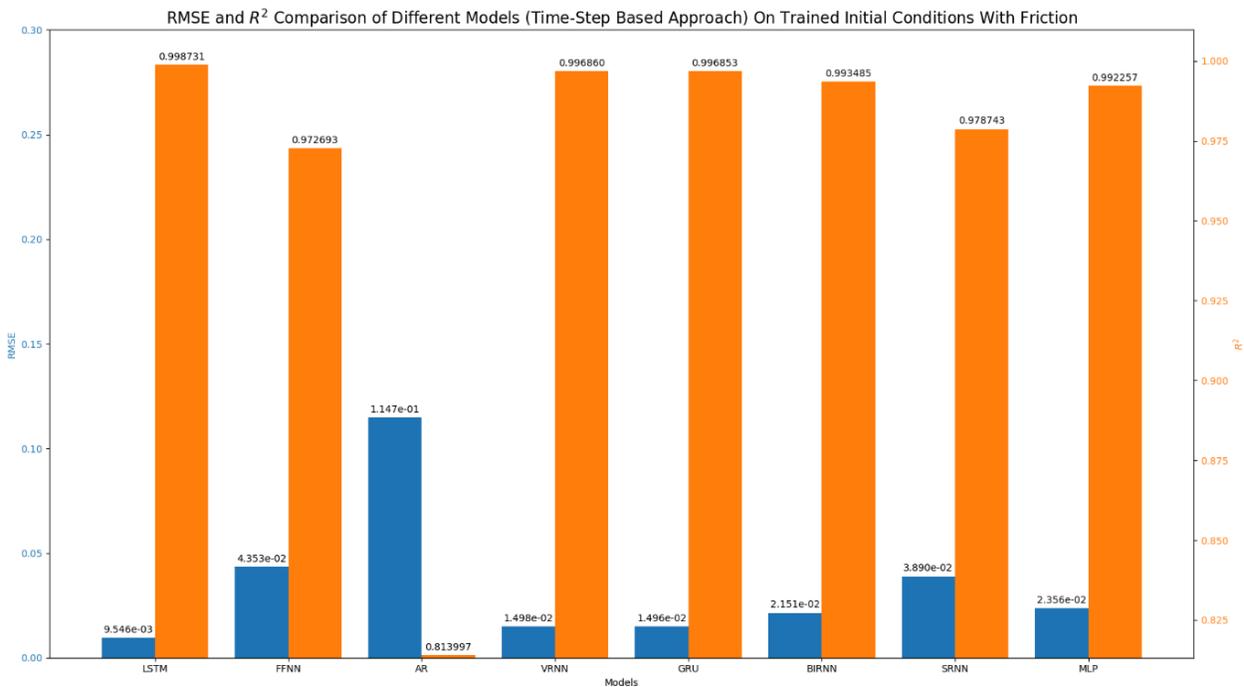

**Fig. 12: Double Pendulum Friction Model Results Time Step** Performances on various ML and NN algorithms on the dataset for timestep based approach. Evaluation metrics used were $R^2$ and RMSE for initial angle [120°, 0°]

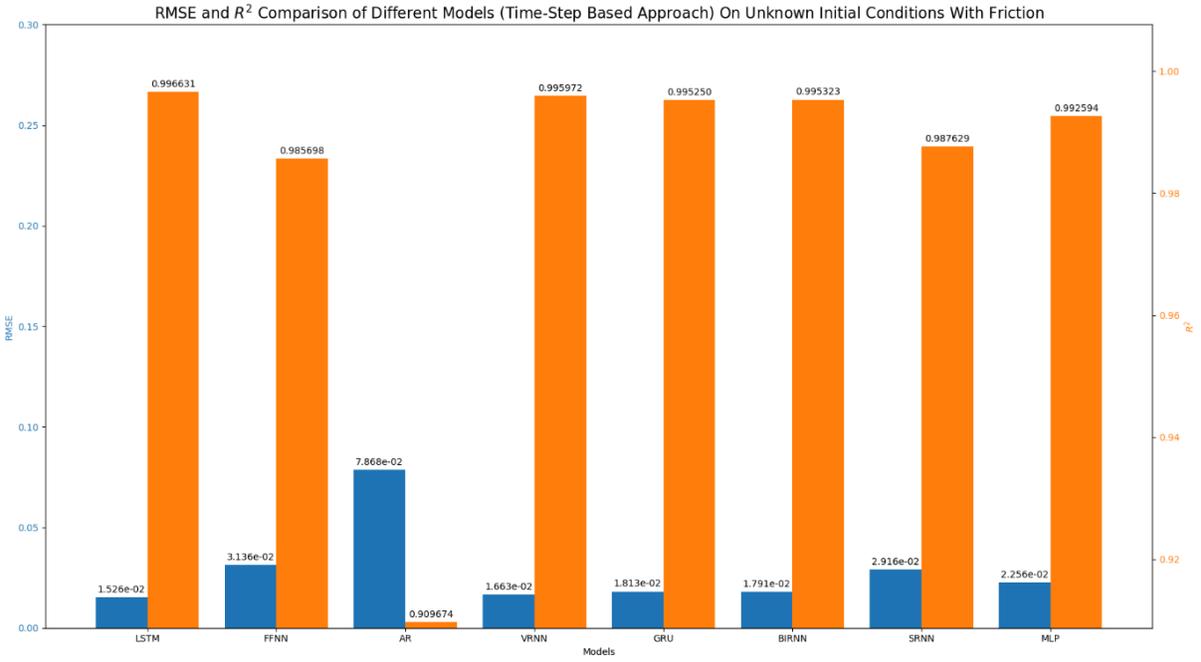

**Fig 13. Double Pendulum Friction Model Results Time Step Based Approach - Testing on unknown initial angle** Performances on various ML and NN algorithms on the dataset. Evaluation metrics used were $R^2$ and RMSE for testing on unknown initial angle [120°, 2.05°]

14A:

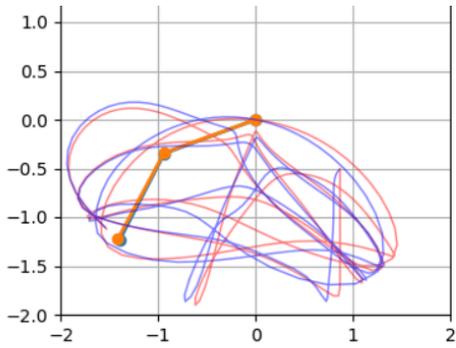

14B:

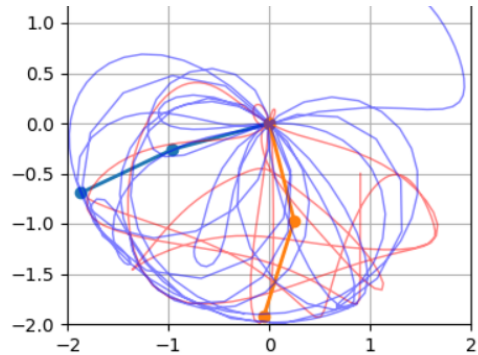

**Figure 14A, 14B; LSTM Trajectory WITHOUT FRICTION for Time Step Based Approach: 10s of trajectory.** Figure A is the trajectory for the model that was trained on many different initial angles and tested on an initial angle that it was already trained on [120°, 0°]. Figure B is the trajectory for the model that was trained on many different initial angles and tested on an "in-between" of [120°, 2.05°]. In both trajectory visualizations, red is the actual trajectory and blue is the predicted trajectory.

15A:

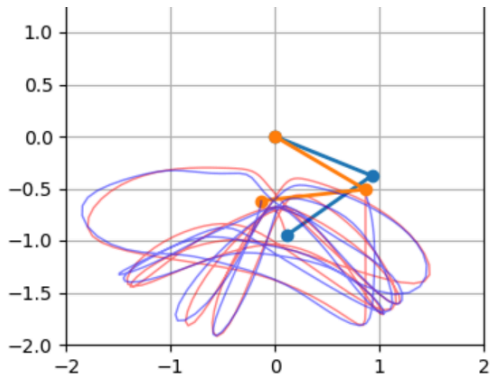

15B:

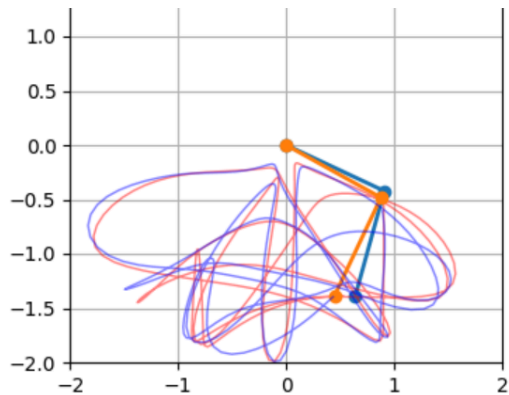

**Figure 15A, 15B; LSTM Trajectory WITH FRICTION for Time step Based Approach: 10s of trajectory**. Figure A is the trajectory for the model that was trained on many different initial angles and tested on an initial angle that it was already trained on [120°, 0°]. Figure B is the trajectory for the model that was trained on many different initial angles and tested on an "in-between" of [120°, 2.05°]. In both trajectory visualizations, red is the actual trajectory and blue is the predicted trajectory.

16A:

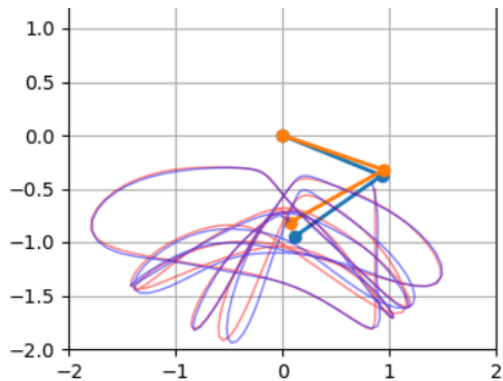

16B:

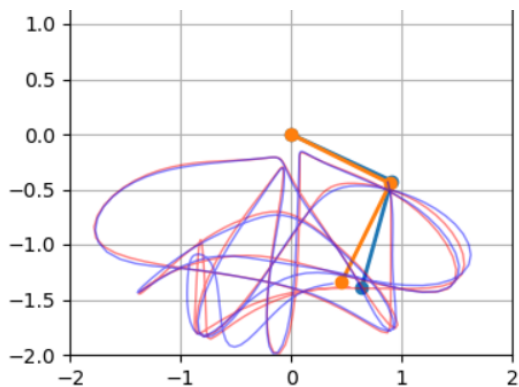

**Figure 16A, 16B; VRNN Trajectory WITH FRICTION for Time step Based Approach: 10s of trajectory.** Figure A is the trajectory for the model that was trained on many different initial angles and tested on an initial angle that it was not trained on [120°, 2.05°]. Figure B is the trajectory for the model that was trained on many different initial angles and tested on one of its trained angles [120°, 0°]. In both figures, the orange is the actual trajectory and blue is the predicted trajectory.

When tested with the sliding window approach at an initial angle [90°, 90°], the best-performing model overall was LSTM with the lowest RMSE value of $5.75_E{}^{-5}$ {Figure 10A}. This was followed by GRU with an RMSE of $6.2_E{}^{-4}$ {Figure 10A}. The worst models were RF and AR with RMSE values of $1.5_E{}^{-2}$ and $2.9_E{}^{-2}$, respectively {Figure 10A}. Typically, all the neural networks performed better than all of the simple machine learning models, which is not a surprise as Recurrent Neural Networks are known to be best suited for time-series problems such as chaotic mechanical systems.

     However, once we realized the limited capabilities of the sliding window approach in modeling chaos, we switched to the new time-step based approach, where we actually got similar results.

     The multiplicative increased data wrought from each set of initial angles caused decreased accuracy for specific pre-trained initial angles than for the sliding window approach. However, the second implementation allowed us to predict in-between unknown angles.

First, testing on an angle that was trained on ([120°, 0°]) without friction, the best model was LSTM again with an RMSE of $2.7_E{}^{-2}$ and an $R^2$ value of 0.991 {Figure 11}, beating all the models with superior performance and best trajectory match between predicted and actual {Figure 14A}. GRU and VRNN also came close to LSTM in terms of results, with an RMSE of $3.84_E{}^{-2}$ and $4.02_E{}^{-2}$, respectively.

These results, although promising, are not necessarily modeling real world chaotic systems. This prompted a change, shifting to friction based systems. This was mainly driven by the need to stabilize chaotic trajectories and improve model performance on unseen conditions. Friction inherently dampens chaotic motion, reducing sensitivity to small perturbations in initial conditions. This allowed models to learn patterns that it might not have caught in a frictionless system, due to its extreme variability. In addition, most real world chaotic systems have damping or dilation of chaos over time: weather catastrophes, medicinal effects, and even the stock market with bigger companies. So including friction in our paper makes the results amenable to the physics oriented scope and other real world chaotic systems.

With friction applied, LSTM was the best model at an initial angle [120°, 0°] with an RMSE of $9.5_E{}^{-3}$ and an $R^2$ value of 0.9981, beating the GRU by a small margin ($1.496_E{}^{-2}$ RMSE; 0.9968 $R^2$ score). The trajectories between the predicted pendulum and actual pendulum are almost indistinguishable for LSTM on a friction-based system {Figure 15A}. This tells us that LSTM is the best model at predicting chaotic systems with two variables, followed by GRU and VRNN. The worst model was AR for a friction-based system, with an RMSE of $1.14_E{}^{-1}$, which suggests that it is not effective in predicting chaotic systems with two variables. A greater number of variables would increase chaos in the system, so AR is not a suitable model to predict chaotic motion.

The key aspect of the second approach, and the basis of our results, was training on multiple different initial angles to see the capabilities of the model. The aim was to predict on datasets with completely different initial angles, changing patterns and other temporal dependencies that were captured by the model for the angle it was trained on. Since these systems are so sensitive to initial conditions that they are considered "unpredictable," evaluating the performance of our models in this manner thoroughly models chaotic systems as a whole. We first tried trying to do this without friction. When tested on [120°, 2.05°], the results were not desirable for frictionless models {Figure 14B} as shown by the big difference in the trajectories of the pendulum for the first 10 seconds of motion for the LSTM, our best model. Without friction, pendulums are extremely volatile in changing conditions. The results were dramatically worse; the LSTM returned an RMSE of 0.26 and an $R^2$ of 0.23 on the unknown initial angle. Because the frictionless pendulum was so chaotic, we decided that it was not worth continuing to test new models after our best model without friction had done terribly without friction. To accommodate for this, we had to shift to modeling pendulums with friction to understand the capability of the models better in prediction on unknown angles, following the justification for damping in real-world scenarios. For friction, the models performed very differently to how the models performed on a frictionless system, with the LSTM returning an RMSE of $1.5_E{}^{-2}$ and an

$R^2$ of 0.996 on unknown angle [120°, 2.05°]. VRNN was one of the best model on a friction-based system with an already trained angle, so it also performed similar to the LSTM in this friction system on the new initial angle, with an RMSE of $1.6_E{}^{-2}$ and an $R^2$ of 0.9959 on unknown angle [120°, 2.05°]. These results were highly desirable, and we can see that the trajectories of the actual vs. predicted closely line up in Figure 15B as it did in Figure 15A. This reveals that LSTM and VRNN are useful even for predicting chaotic situations that they have not always been presented before, but only in reasonably chaotic scenarios (friction-based systems) with two variables as opposed to highly chaotic simulations.

### 3.3 Triple Pendulum

For the triple pendulum, we similarly found that different initial angles exhibited different levels of chaos when disregarding outside factors such as friction or air resistance. When $[\theta_1, \theta_2, \theta_3] \geq [94.737°, 39.131, 33.333°]$, the pendulum demonstrated extremely chaotic behavior. When $[\theta_1, \theta_2, \theta_3] \leq [64.285°, 72°, 91.818°]$, the pendulum demonstrates stability, almost resembling the trajectory of a single pendulum. Similar to the double pendulum, we hypothesized that an increase in $\theta_1$ has a greater correlation with chaos in the pendulum than $\theta_2$, which has a greater correlation with chaos than $\theta_3$.

17A:

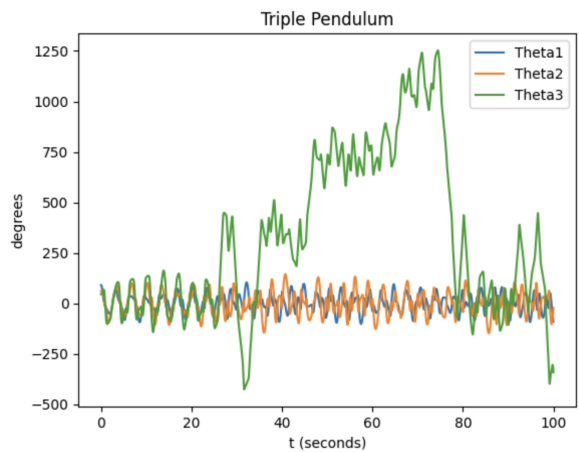

17B:

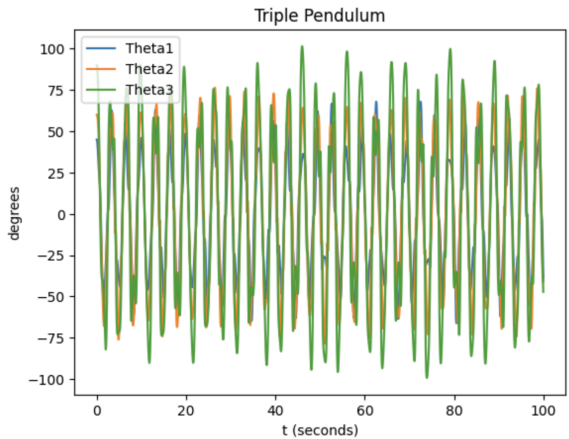

**Figure 17A, 17B; Chaotic vs. Non-Chaotic Motion**; Figure 17A depicts the graph with initial angles [$\theta_1, \theta_2, \theta_3$] = [90°, 60°, 45°]; Figure 17B depicts the graph with initial angles [$\theta_1, \theta_2, \theta_3$] = [45°, 60, 90°]. These graphs highlight the border point between chaos and stability. Figure A depicts a chaotic trajectory, while Figure B depicts a stable trajectory.

From these graphs, we concluded that training our first models at an initial angle [90°, 60°, 45°] would be suitable for the triple pendulum, as it demonstrated the best balance between chaos and predictability.

18A:

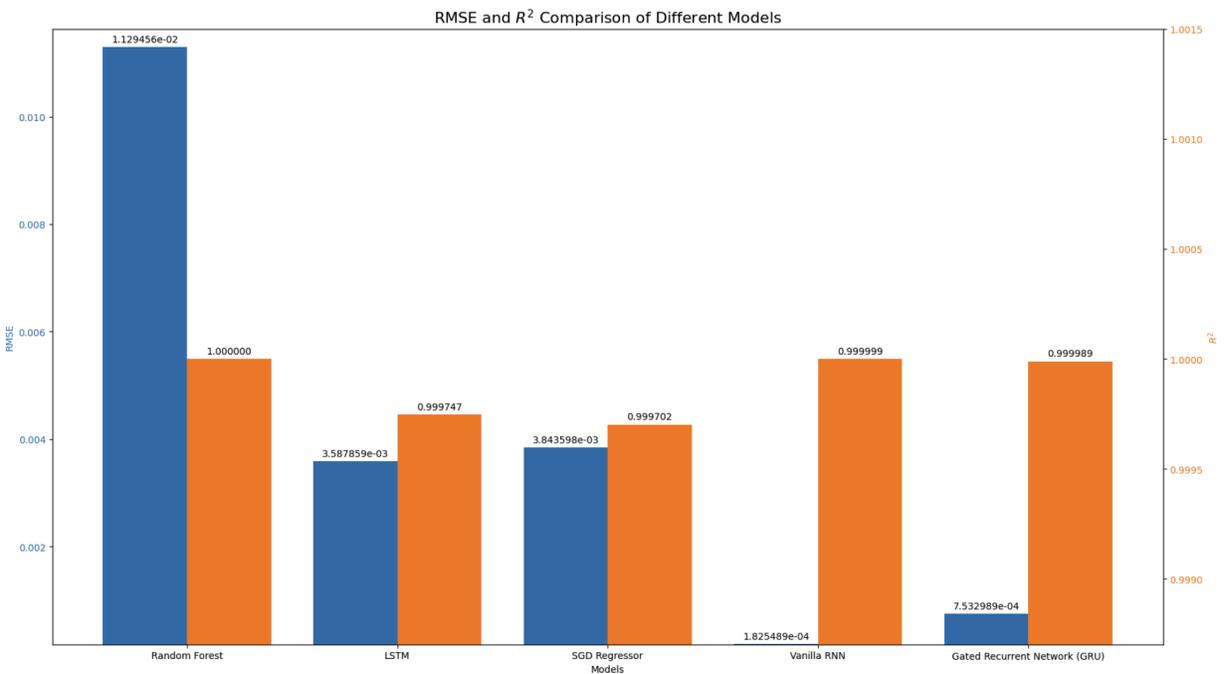

18B:

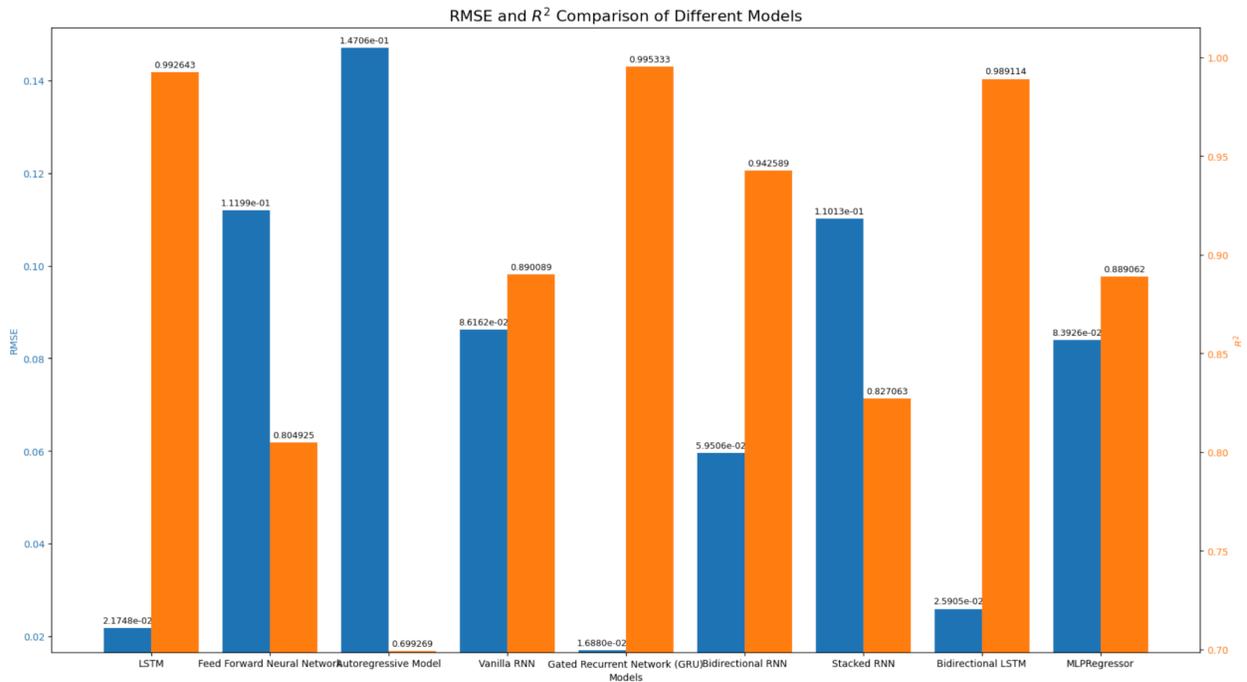

**Fig. 18A, 18B: Triple Pendulum No Friction Model Results (Time-Step Approach)** Performances on various ML and NN algorithms on the dataset. Evaluation metrics used were $R^2$ and RMSE for initial angles [[90°, 90°, 90°], [80°, 80°, 80°] for the old approach (Figure 18A). Following the methodology of the double pendulum, the initial angles for the new approach started at [120°, 0°, 0.1°], and we incremented the third angle by 0.1° until we reached 3.0° (Figure 18B). All results did not take the aspect of friction into consideration.

18C:

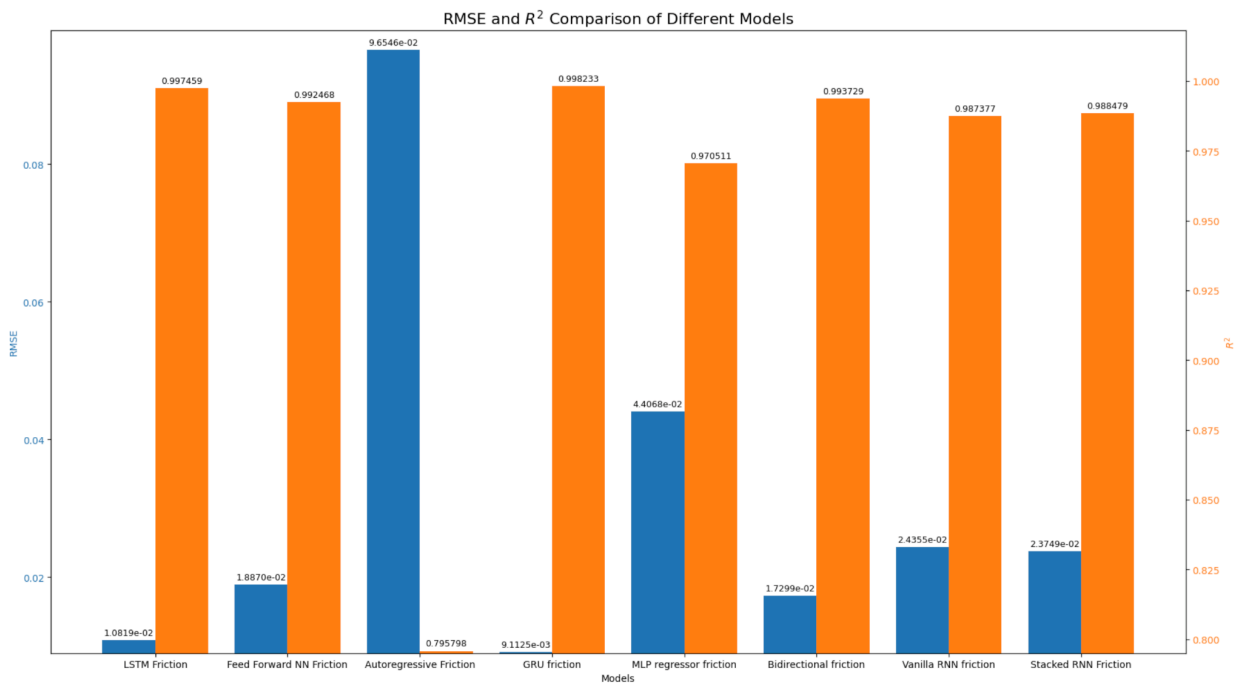

18D:

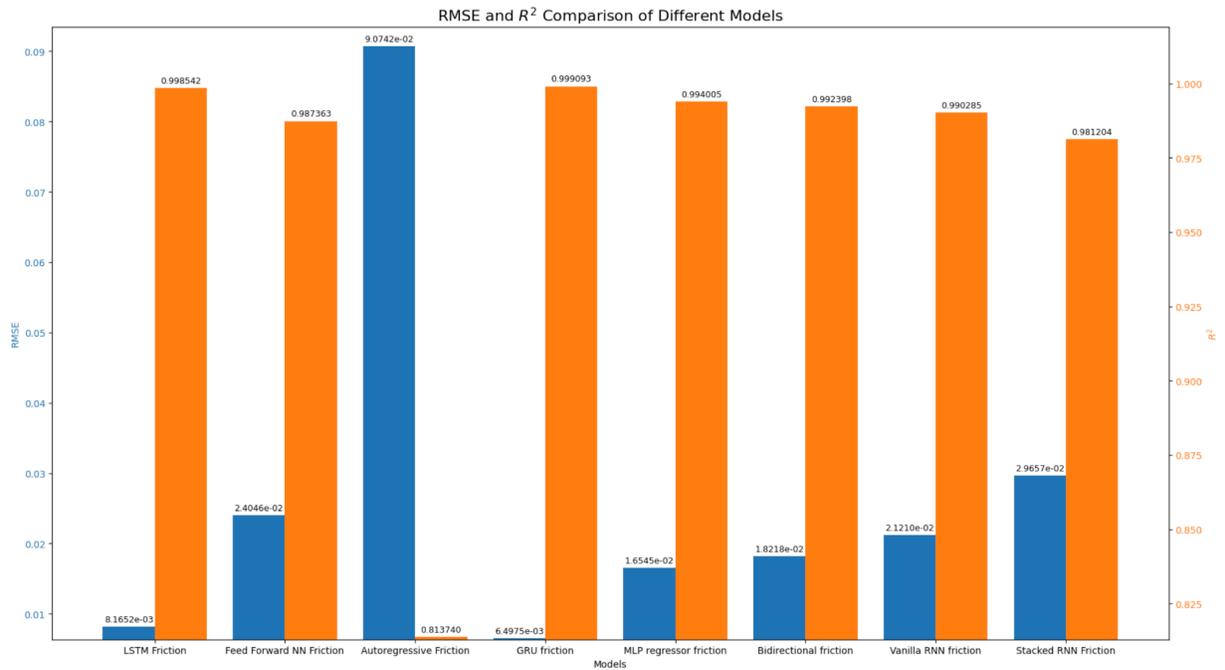

**Fig. 18C, 18D: Triple Pendulum Friction Model Results (Time-Step Approach)**; Performances on various ML and NN algorithms on the dataset. Evaluation metrics used were $R^2$ and RMSE on the trained angles (18C) and on the unknown angle [120°, 0°, 2.05°] (18D). Following the methodology of the double pendulum, the initial angles for the new approach started at [120°, 0°, 0.1°], and we incremented the third angle by 0.1° until we reached 3.0°. All of the results considered the aspect of friction.

19A:

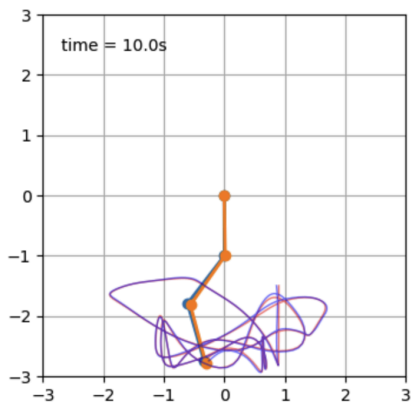

19B:

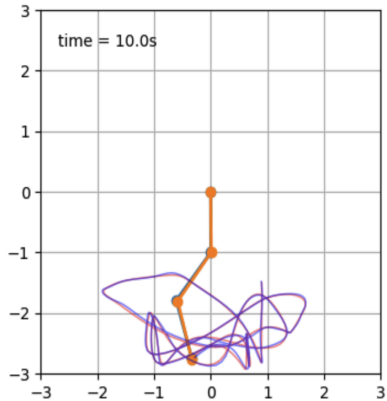

**Figure 19A, 19B; LSTM and GRU Trajectories WITH FRICTION for Time Step Based Approach: 10s of trajectory.** Figure A is the trajectory for the LSTM model that was trained on many different initial angles and tested on an "in-between" angle [120°, 0°, 2.05°]. Figure B is the trajectory for the GRU model that was trained on many different initial angles and tested on an "in-between" of [120°, 0°, 2.05°]. In both trajectory visualizations, red is the actual trajectory and blue is the predicted trajectory.

With the sliding window approach, the Vanilla RNN performed the best out of all the other models, with an $R^2$ score of 0.999999 (perfect data fit), and an RMSE of $1.825_E^{-3}$. We concluded that the Vanilla RNN is the best model to predict the trajectory of the triple pendulum. Contrary to the double pendulum, where the LSTM produced the best results, we hypothesized that the Vanilla RNN captured the motion of the triple pendulum better because the triple pendulum is more sensitive to changes in initial conditions that lead to rapid divergences in trajectories. As the complexity of a chaotic system increases (e.g., moving from a double to a triple pendulum), the ability to predict the system's state over time becomes more difficult. Our data shows that double-pendulum models produced more effective results than triple pendulum results. This increased unpredictability in the triple pendulum's dynamics aligns well with the Vanilla RNN, which can adjust rapidly to local fluctuations.

For our time-step based approach, we trained the models on the angles: $[\theta_1, \theta_2, \theta_3]$ = [120°, 0°, 0.1°], [120°, 0°, 0.2°], … [120°, 0°, 3.0°], varying the third angle throughout the range [0.1, 3.0] with a step of 0.1. With the time-step-based approach, the Gradient Recurrent Network (GRU) performed the best, although not by a lot. With an $R^2$ score of 0.995333 and an RMSE of $1.688_E^{-1}$, the GRU slightly outperformed the LSTM ($R^2$ score of 0.9926 and an RMSE of $2.175_E^{-1}$), using RMSE as our main metric comparison. Based on the GRU's desirable performance, we can conclude that the GRU effectively captures motion given variations in the sets of initial angles in the triple pendulum. These results show that the GRU can generalize well to new, unseen configurations, allowing it to predict the chaotic motion of the triple pendulum.

Now, we moved to test this new approach with friction. To ensure all the models were given the same testing conditions, we decided to test all the models on the angles $[\theta_1, \theta_2, \theta_3]$ =

[120°, 0°, 2.05°]. This ensured that the models were not tested on the trained angle, while still given the opportunity to be trained on the relative "neighborhood" around the angle.

When we tested this new approach with friction, the GRU also performed the best, with an $R^2$ score of 0.98823 and an RMSE of $9.112_E^{-3}$. The LSTM ($R^2$ score of 0.99746 and an RMSE of $1.081_E^{-1}$) and the Bidirectional RNN ($R^2$ score of 0.99372 and an RMSE of $1.73_E^{-1}$) performed well too. We thus conclude that the GRU will best predict the triple pendulum motion when a coefficient like friction is added, diminishing the chaotic motion.

Overall, our results for the triple pendulum – both the old and the new approach – validates our conclusion from the double pendulum. typically, neural networks performed better than simple machine learning models. For the triple pendulum specifically, GRU is known to be best suited in predicting chaotic mechanical systems, both with the time-step approach and the sliding window approach.

## 3.4 Stability Analysis

After coding all our models, we moved on to stability analysis of the pendulum to get a final grasp of its dynamics. At conditions where the Lyapunov exponent was > 0, the system displays chaotic motion as small perturbations in initial conditions are exponentially deviating from the non-perturbed trajectory. {Figure 22}

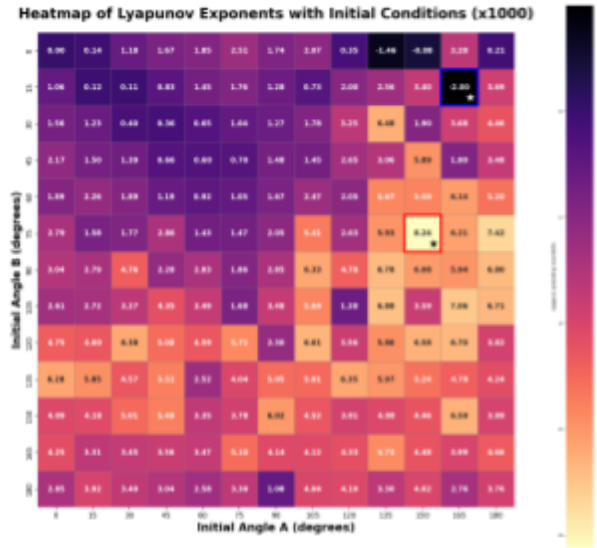

**Fig 20.** Chaoticness at Initial Conditions:
Figure 20 depicts the heatmap which shows the Lyapunov exponent evaluated at various initial conditions for 10 seconds, or 5000 timesteps. The exponent qualitatively assesses the sensitivity to initial conditions: high positive values represent conditions where small deviations grow rapidly, suggesting chaotic behavior, while values close to 0 or negative suggest stability or return to the setpoint. For example, the error of a double pendulum with initial conditions [150.1,75, 0] deviates from one with [150, 75, 0] at a rate of $e^{\frac{8.24}{1000}}$ = 1.008274, or 0.8274% increase after each timestep. This behavior indicates that the system will exponentially move away from the trajectory. On the other hand, the least chaotic point [165.1, 15, 0] deviates at a rate of 0.998, or a -0.1998% increase per timestep, showing that the point tends towards the setpoint's trajectory over time.

In order to assess the stability of the double pendulum system, we found the eigenvalues[40] for the Jacobian matrix of the differential equations at the equilibrium points, namely [0, 0] and [180°, 180°]. At [0, 0], the eigenvalues were [0+5.7873513j, 0-5.7873513j, 0+2.3971994j, 0-2.3971994j], indicating that this point is an undamped oscillator based on Daniel Katzman et al. [10]. Similar analysis at [180, 180] gives [-5.7873513, -2.3971994, 5.7873513, 2.3971994], indicative of an unstable saddle point [10]. This behavior can be further seen with its phase portrait and response vs. time. {Figure 23}

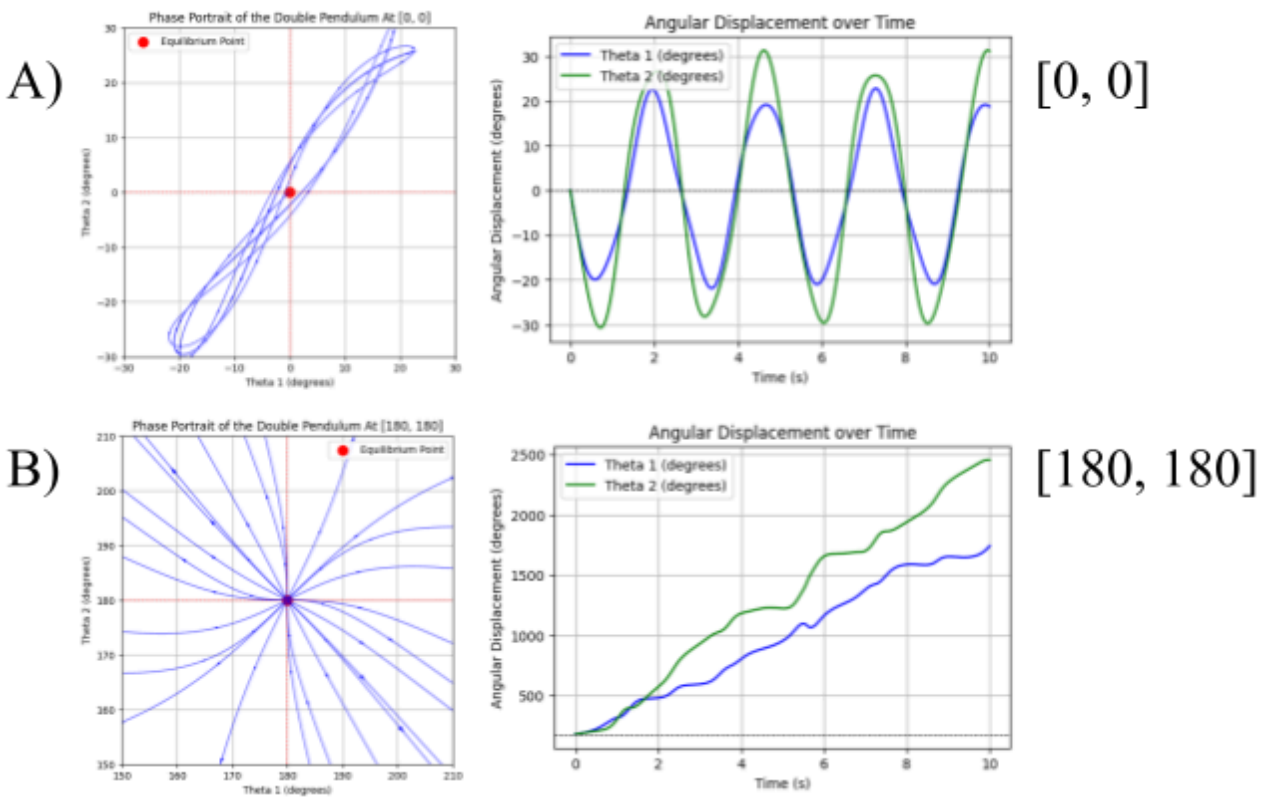

**Fig 21A, 21B. Stability of Equilibrium Points**:
Figure 21A depicts the phase portrait and angular displacement over time for [0, 0] after an initial perturbation of $u_1$ and $u_2$ of ±1 degree/s; Figure 21B depicts the phase portrait and angular displacement over time for [180°, 180°] with a similar perturbation. These graphs highlight how variations in the initial conditions from stable and unstable equilibrium points can lead to divergent results. That is either returning to the setpoint or not reaching the setpoint again, shows the chaotic nature of the system.

Understanding the stability of the multi-pendulum system allows us to assess the reliability of our models as stable initial conditions are able to be predicted within a reasonable error for longer than unstable ones. Additionally, through the Jacobian matrix in the eigenvalue stability, we were able to highlight how parameters affect one another and the overall stability of the system using the partial derivatives of each variable ($[\theta_1, \omega_1, \theta_2, \omega_2]$).

## 3.5 Discussion

Chaotic systems shape the world around us, from climate change and earthquakes to disease spread, financial markets, and urban development. These networks exhibit extreme sensitivity to initial conditions, where a small change can lead to drastically different outcomes. Chaotic systems can be predicted through many computational methods: using differential equation modeling and solving using integration techniques, simulations and real-time data, etc. However, these methods are often extremely time-consuming and inefficient. By harnessing the power of machine learning and deep learning, we can expedite the process of being able to predict these scenarios, and apply it to many fields. In this project, we successfully prove the power of machine learning models in predicting these chaotic systems accurately and efficiently, revolutionizing the field. In addition, we created a novel method of solving the problem efficiently and thoroughly: the time-step based approach, one of the most significant contributions of this paper. Moreover, our works focus on friction models that model chaotic systems in a realistic pattern of damping over time.

Overall, our results from this project suggest that LSTM was the best model to successfully predict chaotic systems with up to 3 features, even ones it hasn't seen before in moderately chaotic scenarios. This can be attributed to LSTM's gated architecture, which allows it to retain critical temporal information while ignoring irrelevant data, making it well-suited for modeling the sensitive dependencies inherent in chaotic systems. In all tests, AR underperformed the most out of the models, which can be attributed to its reliance on linear dependencies, making it inadequate in predicting the non-linear dynamics of a chaotic system. In a broader perspective, most Recurrent Neural Networks like LSTM, VRNN and BIRNN and Convolutional Neural Networks like GRU performed exceptionally, because of their ability to retain time series information, making them useful in predicting other chaotic systems in addition to the double pendulum. Feed Forward Networks like the Autoregressive Model and FFNN underperformed due to their linear dependencies, showing their inability to accurately predict chaotic systems.

For future studies involving investigating machine learning in chaotic systems, we can claim that the majority of systems with two or three variable features using the LSTM model for predictions will achieve an RMSE anywhere between 0.009 and 0.1 for any tests with the right preprocessing. This will revolutionize previously time-consuming algorithms, allowing for the discovery of further discovery of machine learning applications.

While our study demonstrates the capability of machine learning models in predicting chaotic systems, the following limitations exist that put into perspective our result findings.

1. Limitation of Scope:
   While our results are significant for proving the power of machine learning in predicting every-day chaotic systems, they are limited based on the number of variables at play. In this study, we can claim that AI and machine learning will accurately predict chaotic systems with up to 3 changing features, as we were able to predict both the double and triple pendulum accurately. However, adding more features could drastically decrease accuracy, but better data normalization for certain systems could improve the accuracy.

2. Computational Constraints:
   Given the limited computational resources, we narrowed the time-step approach down to a 10-second prediction interval. While this gave ample data for modeling short-term chaos, longer intervals may have been useful in uncovering more dynamics and providing better generalization estimates of the model. Future research would harness more computational power to mitigate any bias due to this.

3. Model Biased to Noise-Free Data:
   The current work has relied purely on simulated data using the solution of the equations of motion via ODE-RK4. This synthetic data is generated in a perfect, idealized environment with RK4 without any noise. Real chaotic systems are susceptible to noise, various kinds of perturbations, and measurement inaccuracies. Future research would include testing the performance of the models on noisy data to better relate model predictions to real chaotic conditions and scenarios. This research would also include getting the equipment to track data for a physical multi-pendulum system, increasing the robustness of our models and reducing possible artifactual results due to simulated data.

4. Uniform Hyperparameters:
   To ensure fairness and keep the results of our models comparable, all of our models shared identical hyperparameters. However, doing this may result in the best models not achieving high performance. Most models, in particular the LSTM and GRU architectures, always require intensive tuning over learning rates, dropout rates, and sizes of hidden layers to achieve optimal performance. The subsequent studies shall model-specific hyperparameter optimizations in order to achieve a more generalized model with increased accuracy.

5. Many Assumed Constants in The Datasets:

There are some assumptions made in the generation of our datasets, and the data may not represent the full spectrum of chaotic systems found in the real world. The variation in the system parameters such as mass or gravitational fields would be a goal for future research on this topic.

By overcoming the above limitations, future studies could extend our work toward more robust and general models applicable for chaos systems. Based off of our research, some specific models stick out

## 4. CONCLUSION

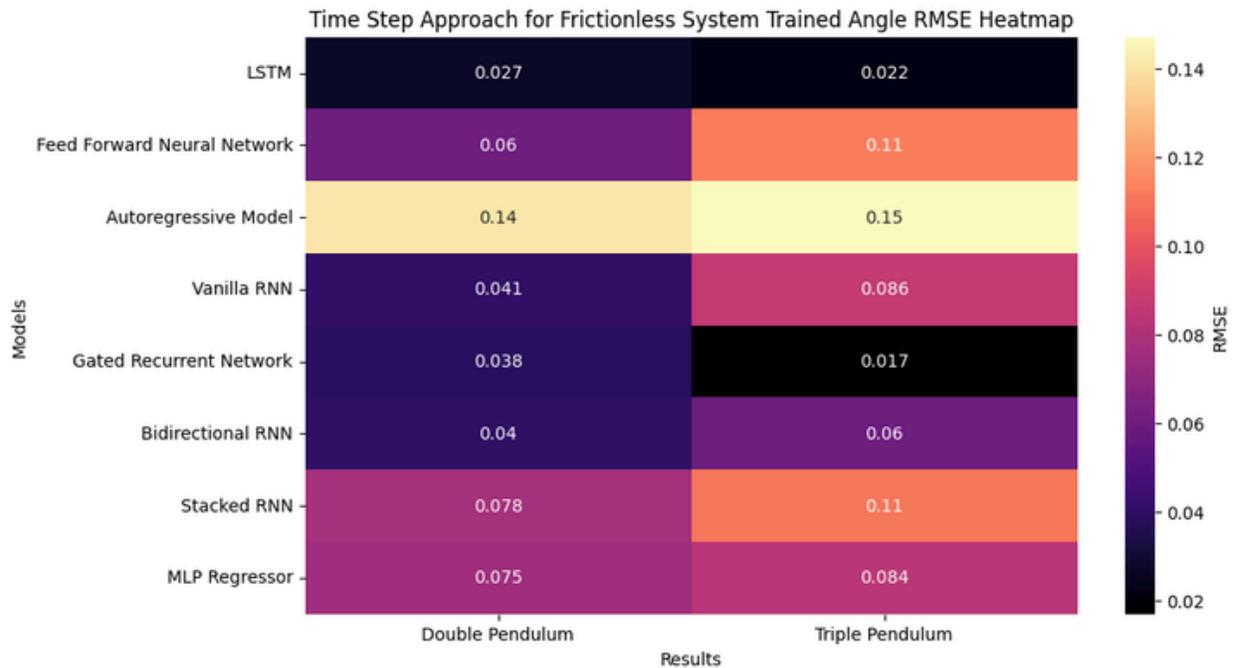

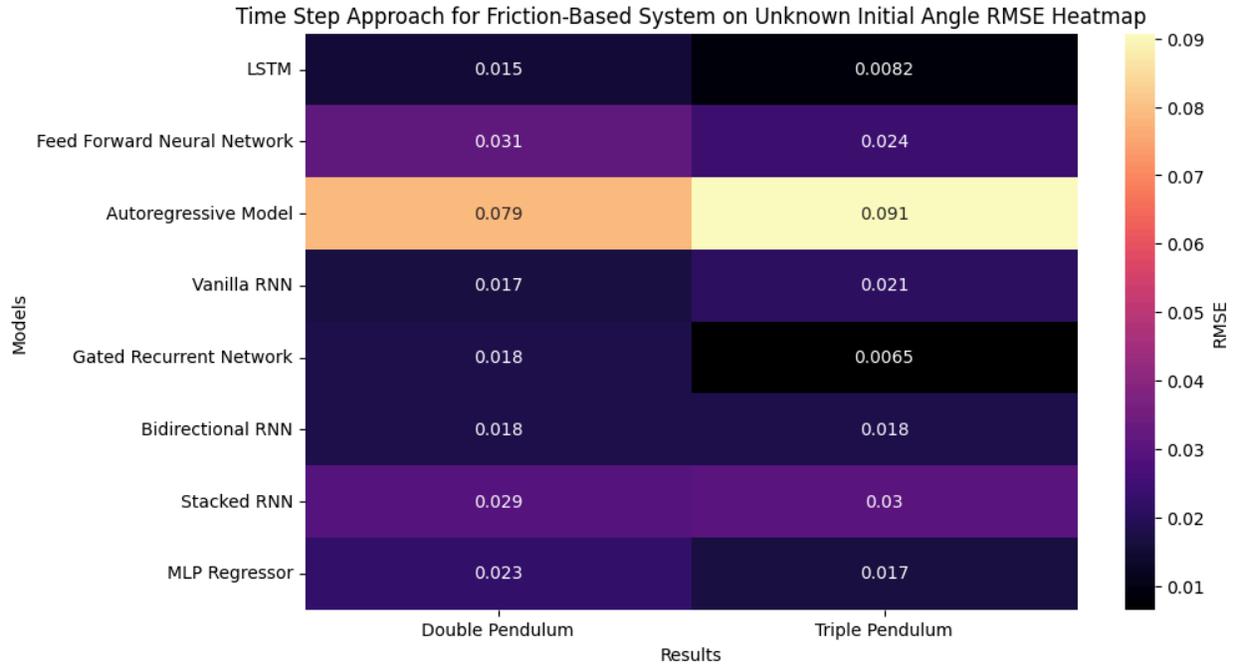

**Fig. 22:** Heatmaps of all Root Mean Square Error (RMSE) values for the models and networks coded with the new method. The lighter the box, the greater the RMSE and the darker the box, the lower the RMSE.

Our study aimed to showcase the power of machine learning techniques in predicting chaotic and almost "unpredictable" systems, specifically the multi-pendulum system. For double pendulum, with our sliding window approach, LSTM had the highest $R^2$ of 0.999998 and an RMSE of $5.75_E^{-5}$. However, the sliding window approach had significant limitations when applied to chaotic systems. It relied on short-term data dependencies, which are insufficient and impractical for modeling the long-term evolution of chaotic trajectories. Consequently, the models, including the LSTM, were merely fitting an erratic curve rather than capturing the underlying chaotic dynamics. This realization prompted the transition to the time-step-based approach. With this new approach for double pendulum, we tested both frictionless and friction-based systems. In a frictionless system on the trained initial angle [120°, 0°], LSTM had the highest $R^2$ of 0.991 and the lowest RMSE of $2.7_E^{-2}$. This was followed closely by GRU with an $R^2$ of 0.982 and an RMSE of $3.84_E^{-2}$. In a friction-based system on the trained angle [120°, 0°], LSTM achieved the highest $R^2$ of 0.9987 and the lowest RMSE of $9.546_E^{-3}$. Following close second was GRU with an $R^2$ of 0.9968 and an RMSE of $1.496_E^{-2}$. When testing on the new chaotic initial angle [120°, 2.05°] without friction, LSTM performed the best with an RMSE of $3.1_E^{-1}$, matching the general patterns of the actual angles of the double pendulum but unable to predict the trajectory accurately as intended due to the unpredictable chaos of the pendulum at this point. When tested on the unknown angle [120°, 2.05°] with friction, LSTM achieved the best result of $1.526_E^{-2}$ and an $R^2$ of 0.9966.

For the triple pendulum, with our sliding window approach, the Vanilla RNN produced the most effective results, with an $R^2$ score of 0.999999 and an RMSE of $1.825_E^{-4}$. With the time-step based approach the GRU had the best results, with an $R^2$ score of 0.995333 and an RMSE of $1.688_E^{-2}$. When we tested the time-step based approach with friction on the untrained angle, the GRU also produced the best results, an $R^2$ score of 0.99909 and an RMSE of $6.497_E^{-3}$. Overall, we conclude that based on the results of the novel time-step based approach, recurrent neural networks like the GRU and the LSTM are successfully able to predict the chaotic motion of the triple pendulum systems.

Further future research directions include applying these models and evaluating their performance on other chaotic systems such as the Lorenz Attractor [43] (typical model of chaos), and more industry based systems such in stock market predictions [44], weather pattern predictions [45], etc. Our research is extremely relevant to the rising interest in Artificial Intelligence (AI) to allow for faster and less intensive computation of complex tasks. Chaotic systems are prevalent in everyday life; being able to accurately predict these systems could revolutionize the field of Artificial Intelligence.

## CRediT authorship contribution statement

**Vasista Ramachandruni:** Conceptualization, Methodology, Validation, Investigation, Formal Analysis, Data Curation, Writing - Original Draft, Visualization. Project Administration. **Sai Hruday Reddy Nara:** Conceptualization, Methodology, Validation, Investigation, Formal Analysis, Data Curation, Writing - Original Draft, Visualization. **Geo Lalu:** Methodology, Data Curation, Validation, Investigation, Formal Analysis, Writing - Original Draft, Visualization. **Sabrina Yang:** Methodology, Validation, Investigation, Formal Analysis, Writing - Original Draft, Visualization. **Mohit Ramesh Kumar:** Methodology, Validation, Investigation, Formal Analysis, Writing - Original Draft, Visualization. **Aarjav Jain:** Methodology, Validation, Investigation, Formal Analysis, Writing - Original Draft, Visualization. **Pratham Mehta**: Validation, Investigation, Formal Analysis, Writing - Original Draft. **Hankyu Koo:** Validation, Investigation, Formal Analysis, Visualization. **Jason Damonte:** Investigation, Formal Analysis, Writing - Original Draft. **Marx Akl:** Supervision, Conceptualization, Resources, Writing - Review and Editing.

## Declaration of Competing Interest

The authors declare that they have no conflicts of interest.

## Glossary

Glossary located in [Electronic Supplementary Information](Electronic Supplementary Information).

# Data Availability

Data is available in the GitHub repositories linked at https://github.com/CTNN-ASDRP.

# Acknowledgments

We thank the Aspiring Scholars Directed Research Program (ASDRP) for providing us with access to a research poster printer for symposiums and research conferences. We would like to thank Ved Vyas for the autoregressive model for double pendulum and Feed Forward NN with friction for double pendulum. We would also like to thank Joshveer Singh for training random forest and LSTM models for triple pendulum and writing summaries of potential sources. We thank Arnav Bansal for his contributions on the early work of the double pendulum friction models and for initial analysis of the double pendulum. We thank Huifang Qin, Professor of Computer Science and Artificial Intelligence at the ASDRP Department of Computer Science and Engineering and Professor Robert Downing, Chair Emeritus of the ASDRP Department of Computer Science and former Professor of Computer Science and Astrophysics and Engineering for reviewing the contents of this manuscript and offering valuable feedback.